\documentclass{article}

\PassOptionsToPackage{numbers, compress}{natbib}

\usepackage[final]{neurips_2025}

\usepackage[utf8]{inputenc} 
\usepackage[T1]{fontenc}    
\usepackage{hyperref}       
\usepackage{url}            
\usepackage{booktabs}       
\usepackage{amsfonts}       
\usepackage{nicefrac}       
\usepackage{microtype}      
\usepackage{xcolor}         
\usepackage{amsmath}
\usepackage{multirow}
\usepackage{wrapfig}
\usepackage{todonotes} 
\usepackage{amsthm}
\usepackage{algorithm}
\usepackage{algorithmic}
\usepackage{subfigure}
\usepackage[capitalize, noabbrev]{cleveref}

\newcommand{\TODO}[1]{}

\title{Learning Sparse Approximate Inverse Preconditioners for Conjugate Gradient Solvers on GPUs}

\author{%
  Zherui Yang\textsuperscript{\textnormal{1}}\quad
  Zhehao Li\textsuperscript{\textnormal{2}}\quad
  Kangbo Lyu\textsuperscript{\textnormal{3,4}}\quad
  Yixuan Li\textsuperscript{\textnormal{1}}\quad
  Tao Du\textsuperscript{\textnormal{3,4}}\quad
  Ligang Liu\textsuperscript{\textnormal{1\S}}\\
  \textsuperscript{\textnormal{1}}University of Science and Technology of China\quad
  \textsuperscript{\textnormal{2}}Stanford University \\
  \textsuperscript{\textnormal{3}}Tsinghua University\quad
  \textsuperscript{\textnormal{4}}Shanghai Qi Zhi Institute
}

\newcommand{\bA}{\mathbf{A}}
\newcommand{\bE}{\mathbf{E}}
\newcommand{\bL}{\mathbf{L}}
\newcommand{\bM}{\mathbf{M}}
\newcommand{\bG}{\mathbf{G}}
\newcommand{\bx}{\mathbf{x}}
\newcommand{\bv}{\mathbf{v}}
\newcommand{\be}{\mathbf{e}}
\newcommand{\bb}{\mathbf{b}}
\newcommand{\RR}{{\mathbb{R}}}

\begin{document}

\maketitle

{\renewcommand{\thefootnote}{\S}
 \footnotetext{Corresponding author: \texttt{lgliu@ustc.edu.cn}}
}

\begin{abstract}
The conjugate gradient solver (CG) is a prevalent method for solving symmetric and positive definite linear systems $\mathbf{Ax} = \mathbf{b}$, where effective preconditioners are crucial for fast convergence. 
Traditional preconditioners rely on prescribed algorithms to offer rigorous theoretical guarantees, while limiting their ability to exploit optimization from data.
Existing learning-based methods often utilize Graph Neural Networks (GNNs) to improve the performance and speed up the construction.
However, their reliance on incomplete factorization leads to significant challenges: the associated triangular solve hinders GPU parallelization in practice, and introduces long-range dependencies which are difficult for GNNs to model.
To address these issues, we propose a learning-based method to generate GPU-friendly preconditioners, particularly using GNNs to construct Sparse Approximate Inverse (SPAI) preconditioners, which avoids triangular solves and requires only two matrix-vector products at each CG step.
The locality of matrix-vector product is  compatible with the local propagation mechanism of GNNs. 
The flexibility of GNNs also allows our approach to be applied in a wide range of scenarios.
Furthermore, we introduce a statistics-based scale-invariant loss function. 
Its design matches CG's property that the convergence rate depends on the condition number, rather than the absolute scale of $\mathbf{A}$, leading to improved performance of the learned preconditioner.
Evaluations on three PDE-derived datasets and one synthetic dataset demonstrate that 
our method outperforms standard preconditioners (Diagonal, IC, and traditional SPAI) and previous learning-based preconditioners on GPUs. We reduce solution time on GPUs by 40\%-53\% (68\%-113\% faster), along with better condition numbers and superior generalization performance.
\end{abstract}

\section{Introduction}
Solving symmetric positive definite(SPD) sparse linear systems $\mathbf{A}\mathbf{x}=\mathbf{b}$ is essential in scientific computing and has wide applications in science and engineering, while the development of efficient numerical solvers remains a challenge. 
The conjugate gradient (CG) method is widely used for its efficiency, yet the convergence rate depends critically on preconditioning techniques that effectively transform the original linear system into an easier one to solve at low computational cost. 

Traditional preconditioners have a long history of development and have achieved considerable success.
Although modern GPUs provide massive parallelism, existing preconditioners struggle to leverage it, becoming the computational bottleneck despite their convergence benefits.
The diagonal preconditioner offers a simple and GPU-friendly implementation but provides limited convergence improvement.
Incomplete Cholesky (IC) demonstrates strong CPU performance and effectiveness in ill-conditioned problems \cite{Lin1999IncompleteCF}. However, its requires two triangular solves at each CG step, which are typically difficult to parallelize \cite{Liu2016ASA, Yamazaki2020PerformancePS}, hindering its performance on GPUs.
Recent learning-based approaches \cite{li2023learning, hausner2024neural, trifonov2025learninglinearalgebragraph} aim to improve traditional approaches using neural networks, particularly factorization-based methods (e.g., IC) and graph neural networks (GNNs). 
By substituting the sequential factorization process with a GNN evaluation, these methods reduce the computational time required for incomplete factorizations.
The trainability of GNNs enables the generation of higher-quality factorizations, thereby accelerating the convergence of iterative solvers.
However, these methods also come with certain limitations.
Prior works inherit the limitation of factorization-based methods that involve triangular solves, restricting their GPU parallelization efficiency. The triangular solves also pose challenges for GNNs to capture long-range dependencies and global information through the elimination tree, while GNNs aggregate information from local neighborhoods and struggle to effectively model such interactions \cite{trifonov2025canmpgnnapproximate}.
Additionally, many existing approaches require computing the solution vector $\bx = \bA ^{-1} \bb$ of the original linear system for every $\bA$ in the dataset to evaluate their loss functions \cite{li2023learning, alexander2024fcgno, trifonov2024learning}. Generating datasets containing large matrices becomes increasingly computationally expensive. The varying scales of $\bA$ further pose difficulties for models in learning to improve the condition number of the preconditioned system.


To address these challenges, we present a novel approach for learning GPU-friendly preconditioners using GNNs. 
Our approach directly approximates the inverse $\bA^{-1}$ with a sparse matrix and formulates its construction as a graph-learning problem, which shares conceptual parallels with the traditional Sparse Approximate Inverse (SPAI) preconditioner \cite{Scott2023AlgorithmsFS, saad2003iterative}.
The sparse matrix-vector product (SpMV) is the only routine required at each CG iteration, allowing our approach to leverage GPU acceleration throughout both the construction and CG solving phases.
We argue that the local nature of SpMV naturally complements the local propagation mechanism of GNNs.
Our approach avoids imposing restrictive triangular structures on the preconditioner, enabling us to preserve the sparsity pattern of $\bA$ or even use a different one.
This design results in improved condition numbers compared to previous SPAI and factorization-based preconditioners.
Furthermore, to eliminate dependence on solution vectors and enable robust training across various datasets, we propose the Scale invariant Aligned Identity Loss (SAI loss). This loss requires only the input matrix $\bA$, and its inherent $\bA$-scale invariance aligns with the scale-invariant convergence behavior of CG solvers, yielding higher performance of learned preconditioners.
Extensive experiments conducted on three PDE-derived datasets and a synthetic dataset validate the effectiveness of our approach compared to previous works. The results show that our approach achieves up to 113\% speedup compared with standard traditional preconditioning techniques and previous learning-based preconditioners. Our approach also achieves a better condition number while maintaining good generalization performance.



In summary, we make the following contributions:
\begin{enumerate}
    \item We propose a learning-based approach to generating GPU-friendly preconditioners, leveraging the GPU parallelism across both the construction and CG solving stages.
    Our approach focuses on generating the SPAI preconditioner using
    GNNs, exploiting the natural alignment between the local computation of SpMV and the local propagation mechanism of GNNs.
    \item We introduce the SAI loss, a novel statistics-based and $\bA$-scale invariant loss function, which reduces datasets' generation costs and enhances the performance of learned preconditioners.
    \item Extensive experiments on three PDE-derived datasets and a synthetic dataset demonstrate that our method achieves lower condition numbers and reduced computation time compared to existing approaches, while also maintaining  generalizability, scalability, and robustness.
\end{enumerate}

\section{Related Works}

\paragraph{Traditional Preconditioners}

Preconditioners $\bM^{-1}$ transform the original linear system $\bA \bx = \bb$ into one easier to solve $\bM^{-1} \bA = \bM^{-1} \bb$ for iterative solvers at low computational cost.
The diagonal preconditioner scales the original system by its diagonal entries \cite{saad2003iterative}, offering a straightforward and GPU-friendly implementation. 
However, it provides limited improvement to the convergence rate.
Incomplete Cholesky (IC) preconditioner \cite{Lin1999IncompleteCF} computes sparse and incomplete triangular factors ($\bM = \bL\bL^\top\approx \bA$). It offers better spectral approximations but suffers from limited parallelism due to its reliance on triangular solves \cite{bjorck2015numerical}. 
Sparse Approximate Inverse (SPAI) preconditioners \cite{Bridson1999OrderingAA, benzi1996spai, kol1993fsai} construct a sparse matrix $\mathbf{M}^{-1} = \bG \bG^\top \approx \mathbf{A}^{-1}$ under sparsity constraints. At each CG step, SPAI use matrix-vector products instead of triangular solves in IC, enabling parallelism suitable for GPU architectures. However, its construction depends on sequential algorithms and often exhibits suboptimal performance.
Alternative techniques, such as polynomial \cite{saad2003iterative}, algebraic multigrid (AMG) \cite{xu2017algebraic}, and domain decomposition \cite{smith1997domain} preconditioners, each balance robustness, memory usage, and parallel efficiency differently.
While effective in their respective domains, they lack generality and often require careful parameter tuning to ensure performance \cite{ranzato2021rlforamg, zou2024autoamg}. Machine learning offers a promising solution to address this limitation.

\paragraph{Learning-based Preconditioners}

Recent research in learning-based preconditioning has progressed across several directions. 
CNNs \cite{lan2024anp}, GNNs \cite{Chen2024GraphNP} and neural operators \cite{alexander2024fcgno}, which directly approximate the action of matrix inversion $\mathrm{NN}(\bA, \bb) \approx \bA^{-1}\bb$, are utilized as preconditioners in many iterative solvers.
Although these approaches can significantly reduce the number of iterations, they involve a full forward pass through the network at each iteration, thereby increasing computational overhead. 
The auxiliary components in neural operators are employed as a subspace in iterative solvers \cite{Luo2024NeuralKI, Kopanickov2024DeepOnetBP, Zhang2022BlendingNO}.
The usage of neural operators restricts their applicability primarily to PDE-related problems, while neglecting more general problems.
\citet{li2023learning, hausner2024neural, trifonov2025learninglinearalgebragraph} have investigated the integration of machine learning and factorization-based preconditioners, specifically by training GNNs to predict the factor $\bL$ in IC preconditioners. However, these methods suffer from the triangular solve due to their sequential computation \cite{Scott2023AlgorithmsFS} and the theoretical challenges for GNNs to model long-range dependencies along the elimination tree \cite{trifonov2025canmpgnnapproximate}.
Although these methods have achieved performance gains compared with the traditional alternatives, few of them consider the performance on 
GPUs, while our approach focuses specifically on GPU performance.

\section{Our Approach}\label{sec:ourapproach}



\subsection{Problem Setup}\label{subsec:prob-setup}
We consider solving the sparse and SPD linear system $\bA\mathbf{x} = \mathbf{b}$, where $\bA \in \RR^{n \times n}$ (or $\RR^{nb \times nb}$ for a blocked sparse matrix with $b$ as block size\footnote{In the hyperelasticity problem, there are 3 variables to solve per node, resulting in a block size of 3.}) arises from diverse applications, such as spatial discretization of PDEs using the finite element method (FEM). 
The matrix $\bA$ is associated with a graph $\mathcal{G}_\bA = (\mathcal{V}, \mathcal{E})$, defined as follows:
\begin{enumerate}
    \item Vertices $\mathcal{V}$: Represent variables (e.g., mesh nodes in FEM) with features $\bv_i \in \RR^d$, that encode geometric or physical properties.
    \item Edges $\mathcal{E} = \{(i,j)\mid \bA_{ij} \ne \mathbf{0} \}$: Connect interacting variables, with edge features $\be_{ij}$ representing the corresponding entries or blocks $\bA_{ij}$.
\end{enumerate}
To accelerate CG's convergence, a preconditioner $\bM^{-1}$ is introduced, which transforms the problem into an easier one to solve by modifying the search direction. The residual vector $\mathbf{r}$ is replaced with the preconditioned direction $\mathbf{s} = \bM^{-1} \mathbf{r}$ at each CG step, as shown in \cref{app:pcg}.
\emph{A good preconditioner $\bM^{-1}$ reduces the condition number $\kappa$ of the linear system by effectively approximating $\bA^{-1}$} and a lower condition number typically indicates faster CG convergence. We consider the following preconditioner construction task based on the graph and its associated features:
\begin{equation}
    (\mathcal{G}_\bA, \{\mathbf{v}_i\}, \{\mathbf{e}_{ij}\}) \xrightarrow{\text{Algorithm}} \bM^{-1} \in \RR^{n \times n},
\end{equation}
where "Algorithm" denotes both traditional approaches (e.g., IC, SPAI, AMG) and neural networks. Traditional approaches construct $\bM^{-1}$ through prescribed and heuristic rules based on the graph structure and its features, while the learning-based approaches infer the mapping from data.



\begin{figure}
    \centering
    \includegraphics[width=1\linewidth]{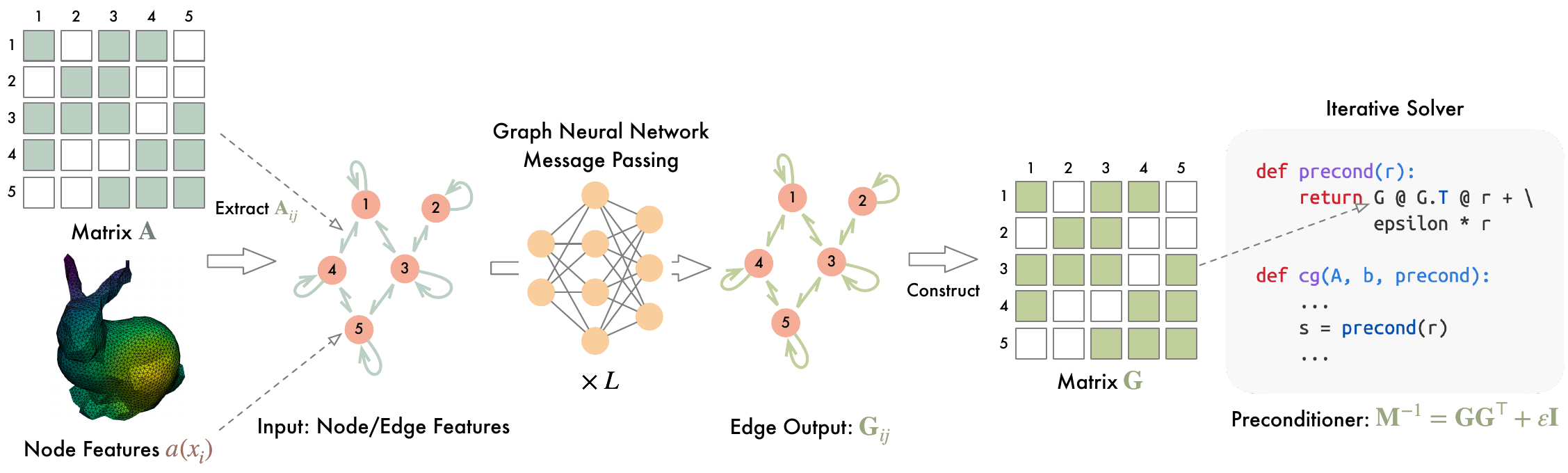}
    \caption{Overview of our approach: By inputting the matrix's nonzero entries $\bA_{ij}$ and node features $a(x_i)$, the GNN processes these features through message passing, and outputs the entries of $\bG_{ij}$. The sparse matrix $\bG$ is assembled and then applied in the preconditioned CG solver.}
    \label{fig:pipeline}
\end{figure}

\subsection{Graph Neural Networks for SPAI construction}\label{subsec:gnn-core}

\subsubsection{Sparse Approximate Inverse Preconditioners (SPAI)}

To ensure high computational efficiency, particularly on GPUs, we directly employ a sparse approximate inverse $\bM^{-1}$ as the preconditioner for $\bA^{-1}$ (SPAI).
Additionally, since the convergence of CG depends on a symmetric and positive definite $\bM^{-1}$, we propose to factorize the preconditioner as
\begin{equation}\label{eq:major}
    \bA^{-1} \approx \bM^{-1} = \bG\bG^\top + \varepsilon \mathbf{I},
\end{equation}
where the sparse matrix $\bG$ is the output of the GNN and $\varepsilon$ is a small positive constant to enforce SPD property, as shown in \cref{fig:pipeline}.
Numerous prior studies \cite{kol1993fsai, benzi1996spai, Bridson1999OrderingAA} have demonstrated that sparse $\bM^{-1}$ can effectively reduce the condition number, even when $\bA^{-1}$ is dense. 
As a result, at each CG iteration, the application of preconditioner $\mathbf{s} = \bM^{-1}\mathbf{r}$ is instantiated as two sparse matrix-vector products and a vector addition, which is both efficient on GPUs.

Previous works \cite{li2023learning, hausner2024neural} take an opposite approach to ours.
Their approaches output a sparse triangular matrix $\bL$, which aim to approximate $\bA$ with $\mathbf{LL}^\top$, resulting in Incomplete Cholesky (IC) preconditioner $\mathbf{M}^{-1} = (\mathbf{LL}^{\top})^{-1}$.
The application of preconditioner $\bL\bL^{\top} \mathbf{s} = \mathbf{r}$ involves two triangular solves: forward substitution $\bL \mathbf{y} = \mathbf{r}$ and backward substitution $\bL^{\top} \mathbf{s} = \mathbf{y}$.
Although triangular solves can be efficient on CPUs, the inherent sequential nature limits their performance on GPUs.

\paragraph{Locality of SPAI Preconditioner}
Consider the application of our preconditioner $\mathbf{s} = \bM^{-1} \mathbf{r}$. Suppose $\bG$ shares the same sparsity pattern as $\bA$, the $j$-th entry of the output vector $\mathbf{s}$ is
\begin{equation}\label{eq:spai-locality}
    \mathbf{s}_j = 
    (\bM^{-1} \mathbf{r})_j = \varepsilon \mathbf{r}_j + \sum_{\substack{l \\ \bA_{jl} \neq 0}} \bG_{jl} \sum_{\substack{k \\ \bA_{kl} \neq 0}} \bG_{kl} \mathbf{r}_k.
\end{equation}
Since the non-zero entries of $\bA$ and the entries of vectors correspond to edges and nodes in the graph, \cref{eq:spai-locality} reveals that \emph{the output on node $j$ depends only on its two-hop neighborhood}, mediated through intermediate nodes $l$ and source nodes $k$. This makes SPAI fundamentally more compatible with GNN architectures than factorization-based methods.

In contrast, in IC preconditioner, the forward substitution $\bL \mathbf{y} = \mathbf{r}$ and backward substitution $\bL^{\top} \mathbf{s} = \mathbf{y}$ propagate information along the elimination tree induced by a chosen node ordering. 
The forward substitution aggregates values from ancestor nodes, while backward substitution collects from descendant nodes, resulting in global dependencies.
This hierarchical propagation in IC conflicts with the local aggregation mechanism of GNNs over fixed-hop neighborhoods. 
Furthermore, IC requires explicit directional dependencies (enforcing lower triangular $\bL$), which are inconsistent with the undirected/symmetric assumptions in existing GNN-based preconditioners.

\subsubsection{GNN Architecture}

Our GNN follows an encoder-processor-decoder architecture, adapted from \citet{gilmerMessagePassing17}. Given node features $\bv_i \in \RR^{d_{\mathrm{node}}}$ and edge features $\be_{ij} \in\RR^{d_{\mathrm{edge}}}$, the GNN first applies two MLPs $E_n, E_e$ to encode the  node and edge features into their hidden representations $\mathbf{x}^{(0)}, \mathbf{h}^{(0)} \in \RR^{d}$:
\begin{equation}
    \mathbf{x}^{(0)}_i = {E}_n(\bv_i), \quad \mathbf{h}_{ij}^{(0)} = {E}_e(\be_{ij}).
\end{equation}
Subsequently, $L$ message-passing layers are applied. Each layer utilizes three MLPs, $f_{m}^{(t)}, f_v^{(t)}, f_e^{(t)}$, which serve as message functions and update functions for node and edge features:
\begin{equation}
\begin{aligned}
    \mathbf{m}^{(t)} &= \sum_{j \in N(i)} f_{m}^{(t)}(\mathbf{x}_{i}^{(t-1)}, \mathbf{x}_{j}^{(t-1)}, \mathbf{e}_{ij}),\\
    \mathbf{x}^{(t)}_{i} &= \mathbf{x}_i^{(t-1)} + f_v^{(t)}(\mathbf{m}^{(t)}_{i}), \quad\
    \mathbf{h}^{(t)}_{ij} = \mathbf{h}_{ij}^{(t-1)} + f_e^{(t)}(\mathbf{x}^{(t)}_{i}, \mathbf{x}^{(t)}_{j}, \mathbf{h}_{ij}^{(t-1)}).
\end{aligned}
\end{equation}

Finally, a decoder MLP ${D}$ is applied to the edge features of the last layer $\mathbf{h}^{(L)}$ to get the block entries: 
\begin{equation}
    \bG_{ij} = {D}(\mathbf{h}^{(L)}_{ij}) \in \RR^{b\times b}.
\end{equation}
The global sparse matrix $\bG\in \RR^{nb\times nb}$ is constructed by replacing each block $\bA_{ij}$ with $\bG_{ij}$ while preserving the original sparsity pattern. 

\subsection{Loss Function}\label{subsec:sailoss}
The loss function is crucial for learning effective preconditioners $\bM$. 
While traditional SPAI approaches aim to minimize $\| \bM^{-1} - \bA^{-1}\|_F$ or $\| \bA\bM^{-1} - \mathbf{I}\|_F$, these formulations suffer from two limitations: 
(1) 
Direct evaluation of $\bA \bM^{-1}$ requires significantly larger memory and computation resources for large matrices even when both $\bA$ and $\bM$ are sparse.
(2) 
These losses depend on the absolute magnitude of $\bA$, while the convergence rate of CG solvers is invariant to the magnitude.

\paragraph{Stochastic Estimation} To address the computational bottleneck, we first adopt the stochastic trace estimator \cite{hausner2024neural, hutchinson1989stochastic} to approximate the matrix Frobenius norm without explicitly constructing $\bA\bM^{-1}$:
\begin{equation}\label{eq:loss-initial}
    \| \bA\bM^{-1} - \mathbf{I} \|_F^2 = \mathrm{tr}((\bA\bM^{-1} - \mathbf{I})^\top (\bA\bM^{-1} - \mathbf{I})) = 	\mathbb{E}_{\mathbf{w}} \left[\| \bA\bM^{-1} \mathbf{w} - \mathbf{w} \|_2^2\right],
\end{equation}
where $\mathbf{w}$ is a vector of independent and identically distributed (i.i.d.) random variables drawn from the standard normal distribution $\mathcal{N}(0, 1)$.
This reduces the computation to matrix-vector products $\bA\bM^{-1} \mathbf{w}$, avoiding costly matrix-matrix products.
The estimator can also be applied to $\| \bM^{-1} - \bA^{-1}\|_F^2\approx \|\bM^{-1}\mathbf{w} - \bA^{-1}\mathbf{w}\|_2^2$. However, this either requires access to $\bA^{-1}$ or involves solving $\bx = \bA^{-1} \mathbf{w}$ for each sample in the dataset, making it computationally expensive 
.

\paragraph{Scale Invariance} The loss function in \eqref{eq:loss-initial} exhibits scale sensitivity because it depends on the absolute magnitude of $\bA$. 
To align with the CG solver's scale invariance property, we normalize $\bA$ by its norm $\|\bA\|$ in the loss function:
\begin{equation}\label{eq:our-loss}  
   \mathcal{L}_{\mathrm{SAI}}(\bA, \bM^{-1}) = \left\| \frac{1}{\|\bA\|} \bA  \bM^{-1} - \mathbf{I} \right\|_F^2, \quad \mathcal{L}_{\mathrm{SAI}}(\bA, \bM^{-1}, \mathbf{w}) = \left\| \left( \frac{1}{\|\bA\|} \bA \bM^{-1} - \mathbf{I} \right) \mathbf{w} \right\|_2^2.
\end{equation}
This formulation ensures $\mathcal{L}_{\mathrm{SAI}}(\bA, \bM^{-1}, \mathbf{w}) = \mathcal{L}_{\mathrm{SAI}}(\alpha\bA, \bM^{-1}, \mathbf{w})$ for any $\alpha>0$. We refer to this as \textbf{Scale invariant Aligned Identity (SAI) loss}, which explicitly decouples the preconditioner learning from the matrix's absolute scale, while enforcing alignment between the scaled preconditioned matrix and the identity matrix. The condition number $\kappa$ of $\bA\bM^{-1}$ can be estimated as
\begin{equation}
    \kappa(\bA\bM^{-1}) = \frac{\sigma_{\max}(\bA\bM^{-1})}{\sigma_{\min}(\bA\bM^{-1})} = \frac{\sigma_{\max}(\mathbf{I} + \bE)}{\sigma_{\min}(\mathbf{I} + \bE)} \leq \frac{1 + \sigma_{\max}(\bE)}{1 - \sigma_{\max}(\bE)} \approx 1 + 2\sigma_{\max}(\bE) = 1 +2 \|\mathbf{E}\|_2,
\end{equation}
where $\bE = \bA \bM^{-1} / \|\bA\|  - \mathbf{I}$ is the error matrix and $\sigma$ denotes matrix singular values (see \cref{app:proof-of-scal} for its detailed proof).
This derivation shows that the SAI loss encourages $\bM^{-1}$ to focus on improving $\bA\bM^{-1}$’s spectral properties, while being invariant to absolute magnitude of $\bA$. 
Although conventional matrix norms of $\bA$ like the Frobenius norm are theoretically valid, their scale is often sensitive to matrix dimensions and outlier entries. We therefore define a more robust and dimension-agnostic norm as $\|\bA\| = \mathrm{mean}_{\bA_{ij}\ne 0} |\bA_{ij}|$.
This setting also ensures that the edge features of the GNN and the loss term remain on a reasonable scale, promoting more stable and efficient optimization across matrices of varying sizes.

\section{Experimental results}\label{sec:experiments}

\begin{figure}
    \centering
    \subfigure[Density plot of examples in TetWild Dataset for Heat/Poisson Equation test case.]{\includegraphics[width=0.4\linewidth]{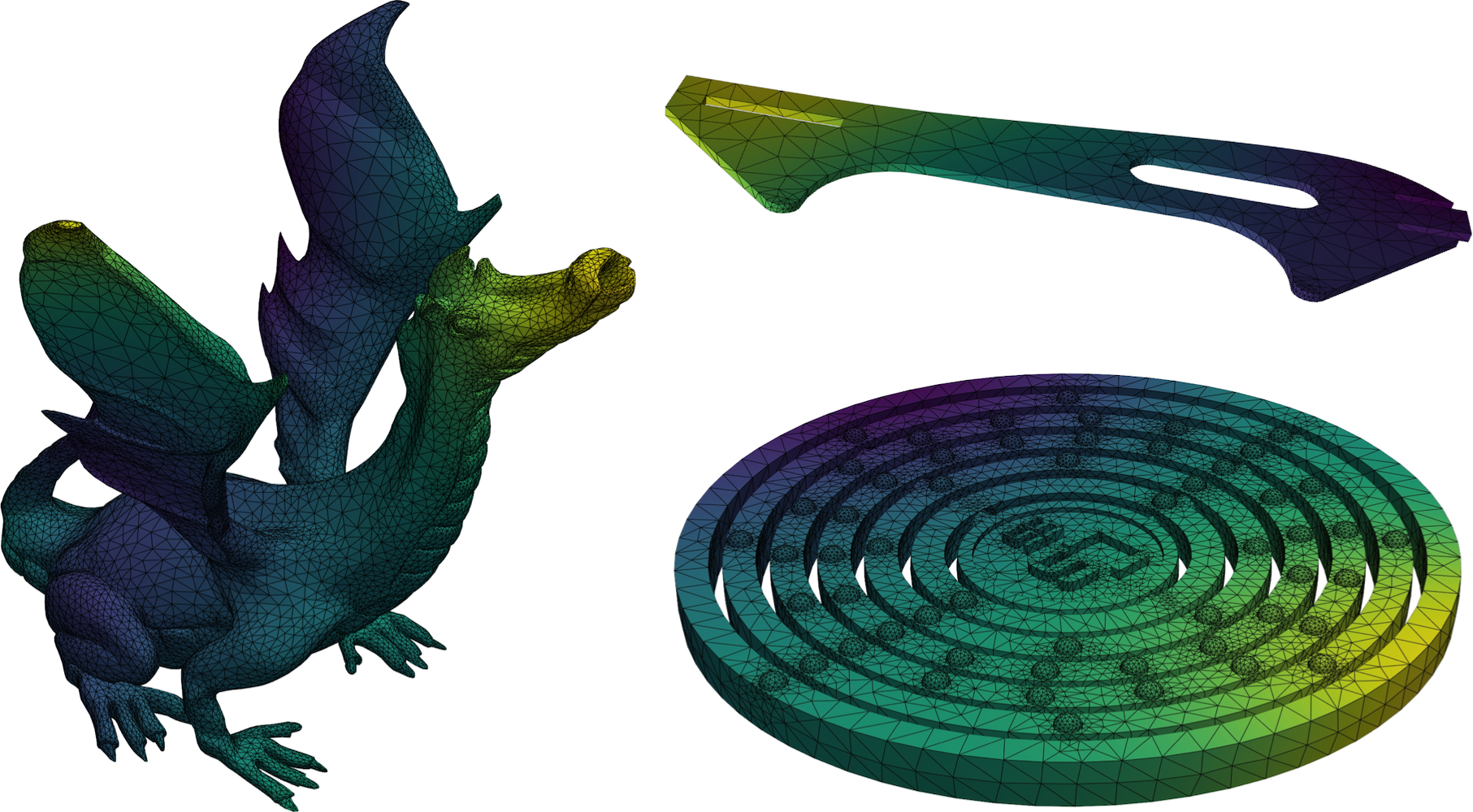}}
    \hspace{1em}
    \subfigure[Starting and ending timestep of meshes with different resolution in the hyperelasticity case.]{\includegraphics[width=0.49\linewidth]{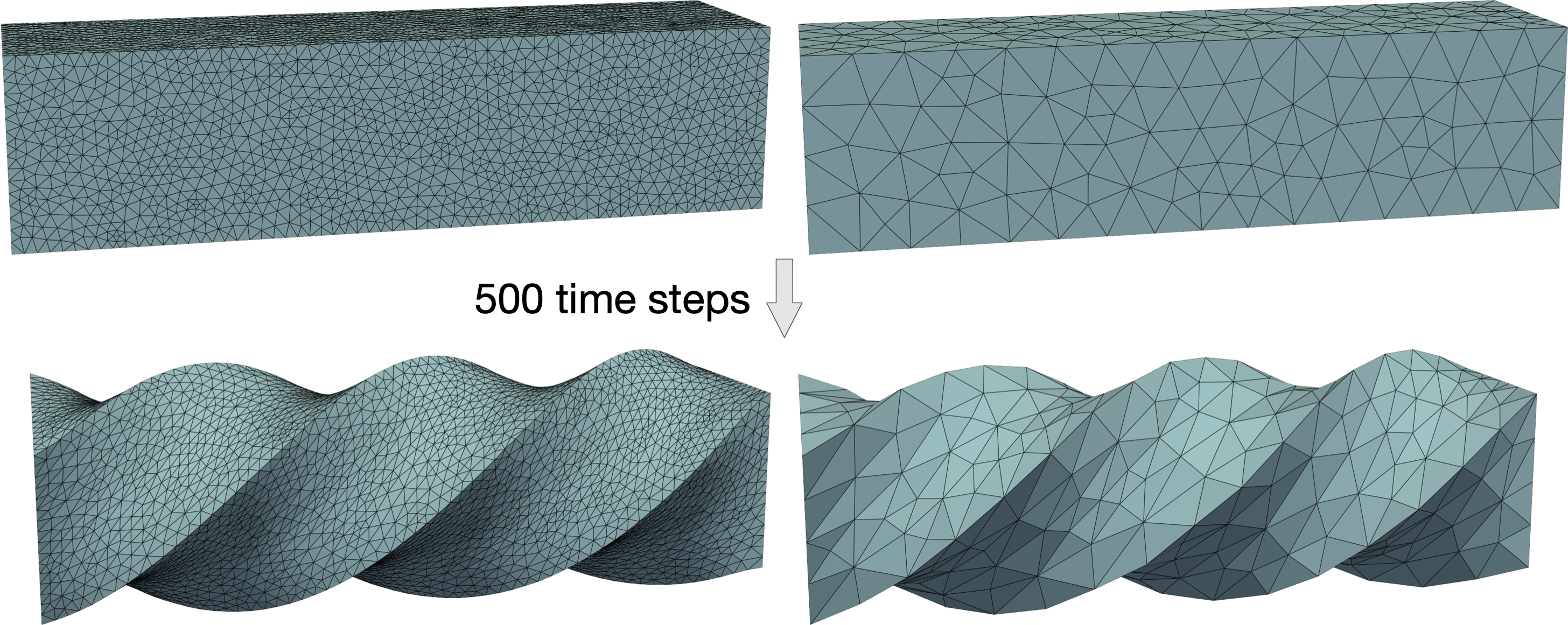}}
    \caption{Examples in our PDE-derived test cases.}
    \label{fig:dataset}
\end{figure}

We answer the following questions through our experiments:
(1) How does our approach compare with traditional and learning-based approaches?
(2) Is our loss more effective than those in prior works?
(3) How do our approach generalize to unseen examples?
(4) Does our approach yield a better condition number compared to previous methods?
We describe the experiment setup in Sec. \ref{subsec:exp-setup} and provide answers to these questions in Sec. \ref{subsec:compare-previous} to \ref{subsec:condition-number}.

\subsection{Experiment Setup}\label{subsec:exp-setup}

Our experiments include three PDE-derived datasets, as well as a fully algebraic dataset composed of randomly generated matrices.
As shown in \cref{fig:dataset}, we use FEM and tetrahedron mesh to discretize the PDEs, yielding the sparse system matrix $\bA$.
The test cases considered in our study are as follows:

\begin{enumerate}
    \item \textbf{Heat Equation}: \( a(x) \cdot u_t - \Delta u = 0 \), where \( a(x) \) represents the spatially varying density (randomly sampled on the mesh, with values ranging from \(1e-4 \) to \( 5e-4 \)).  
    \item \textbf{Poisson Equation}: \( -\Delta u = f \) with \( u|_D = g \), where \( D \) denotes a randomly specified Dirichlet boundary.
    \item \textbf{Hyperelasticity}: \( \mathbf{M}\ddot{\mathbf{u}} + \mathbf{f}_{\mathrm{int}}(\mathbf{u}) = \mathbf{f}_{\mathrm{ext}} \), where \( \mathbf{M} \) is the mass matrix, and \( \mathbf{f}_{\mathrm{int}} \), \( \mathbf{f}_{\mathrm{ext}} \) represent internal and external forces, respectively. The internal force $\mathbf{f}_{\mathrm{int}}$ is nonlinear and derived from a stable Neo-Hookean material model \cite{Kim2022DynamicDI}.
    \item \textbf{Synthetic System}: \( \mathbf{A} = \mathbf{PP}^\top + \varepsilon\mathbf{I} \), where \( \mathbf{P} \) is a random matrix and \( \varepsilon = 10^{-4} \) ensures the numerical positive definiteness of \( \mathbf{A} \).  
\end{enumerate}

For Heat and Poisson problems, we use 9,147 meshes with node counts ranging from 400 to 32,000 from the TetWild dataset \cite{hu2018tetwild}.
All meshes are normalized to ensure the entire geometry fits within the $[-1, 1]^3$ domain.
The hyperelasticity simulations involve a beam twist scene for 500 timesteps. This experiment shares the same geometry but varies in resolution and topology through remeshing with node counts ranging from 645 to 14,039 (matrix size is $3\times$ larger). 
The synthetic dataset contains 1,000 matrices with a sparsity of approximately 0.12\%, with matrix sizes ranging from 10,000 to 20,000. 
All experiments employ a 4:1 train-test split, with all reported results evaluated on the test set. Detailed problem configurations, their matrix representations and corresponding inputs of GNNs are provided in Appendix \ref{app:dataset}.

\paragraph{GNN Implementation and Training} For all experiments in this work, we fixed the number of message passing steps $L$ to 4, the number of hidden layers in all the MLPs to 1, the number of neurons $d$ in the hidden layer to 24, and $\varepsilon = 10^{-4}$. The GNN has about 24k trainable parameters in total. All models are trained for 500 epochs using a batch size of 4 on a single NVIDIA A100 GPU, optimized with AdamW \cite{Loshchilov2017DecoupledWD} and an exponentially decaying learning rate scheduler (decay rate = 0.99). 

\paragraph{Evaluation Metrics} 
The constructed preconditioner $\bM^{-1}$ is subsequently applied in CG to solve $\bA\bx=\bb$. Since each CG iteration requires one preconditioner application, 
the total solving time is governed by:
\begin{equation}
    T_\mathrm{total} = T_\mathrm{construct} + k \times (T_\mathrm{apply} + T_\mathrm{cg}),
\end{equation}
where $k$ denotes iteration count, and $T_\mathrm{construct}$, $T_\mathrm{apply}$ represents the preconditioner's construction time at the beginning and application time per iteration, and $T_\mathrm{cg}$ accounts for fixed operations per CG iteration (e.g., dot products, matrix-vector products). \textbf{The primary objective of preconditioning is to minimize} $ T_{\text{total}} $.
$T_{\mathrm{apply}}$ and $T_{\mathrm{construct}}$ rely on the computational complexity of its associated routines (such as SpMV and triangular solve), while preconditioners with better spectral approximation to $\bA^{-1}$ or smaller condition number often result in lower $k$. This establishes a fundamental trade-off between solver efficiency and approximation quality in preconditioner design.


\subsection{Preconditioner Performance}\label{subsec:compare-previous}

We consider three standard traditional preconditioners on both GPU and CPU architectures, including: (1) Diagonal (Diag), (2) Incomplete Cholesky (IC), and (3) Sparse Approximate Inverse (AINV)\footnote{SPAI is a specific subclass of AINV preconditioners that focuses on constructing a sparse approximate inverse, whereas AINV is the general parent concept of approximate inverse techniques.}. 
All the preconditioned CG are implemented in C++ and CUDA with OpenBLAS \cite{Wang2013AUGEMAG}, cuBLAS, cuSPARSE \cite{Bell2009ImplementingSM}, and cusplibrary \cite{Cusp} for their high-performance linear algebra kernels and preconditioner implementations.
We list their implementation details in Appendix \ref{app:baselines}. 

\paragraph{Comparison to Traditional Approaches}

Table \ref{tab:benchmark} summarizes the time cost $T_{\mathrm{total}}$ and iteration count $k$ for the CG method equipped with different preconditioners to converge to a relative tolerance $\mathrm{rtol}=\|\bb - \bA\bx\| / \|\bb\| < 10^{-8}$. The results demonstrate that our approach outperforms all baseline approaches and achieves consistent superiority across all categories of solving tasks. Additionally, we also present a comparison of our method with AMG in Appendix \ref{app:amg}.

\begin{table}[htbp]
\caption{GPU Benchmark results across different datasets. Total time $T_{\mathrm{total}}$ (ms) and total iterations $k$ (in parentheses) until relative residual norm is less than $\mathrm{rtol}=10^{-8}$ are listed in the table. Lower values indicate better performance. The lowest value is in bold and the second lowest is underlined. The relative promotion (Rel. Prom.) indicates the ratio of the time saved over the second-best one.}
\label{tab:benchmark}
\centering
\begin{tabular}{ccccccc}
\toprule
Test Case & Diag & IC & AINV & Ours & Rel. Prom. ($\uparrow$) \\
\midrule
     Heat Equation& \underline{77}(520) & 167({204}) & 78(330) & \textbf{36}(197) & 113\% \\
Poisson Equation& \underline{45}(320) & 101({128}) & 58(217) & \textbf{26}(128) & 73\% \\
Hyperelasticity& \underline{86}(464) & 202(117) & 247(266) & \textbf{51}({175}) & 68\% \\
Synthetic System& \underline{445}(2775) & 1399({1808}) & 5024(10896) & \textbf{253}(1122) & 75\% \\ 
\bottomrule
\end{tabular}
\end{table}


\begin{figure}[t!]
    \centering
    \subfigure[Total time required to converge for $\mathrm{rtol}=10^{-8}$]{\includegraphics[width=0.52\linewidth]{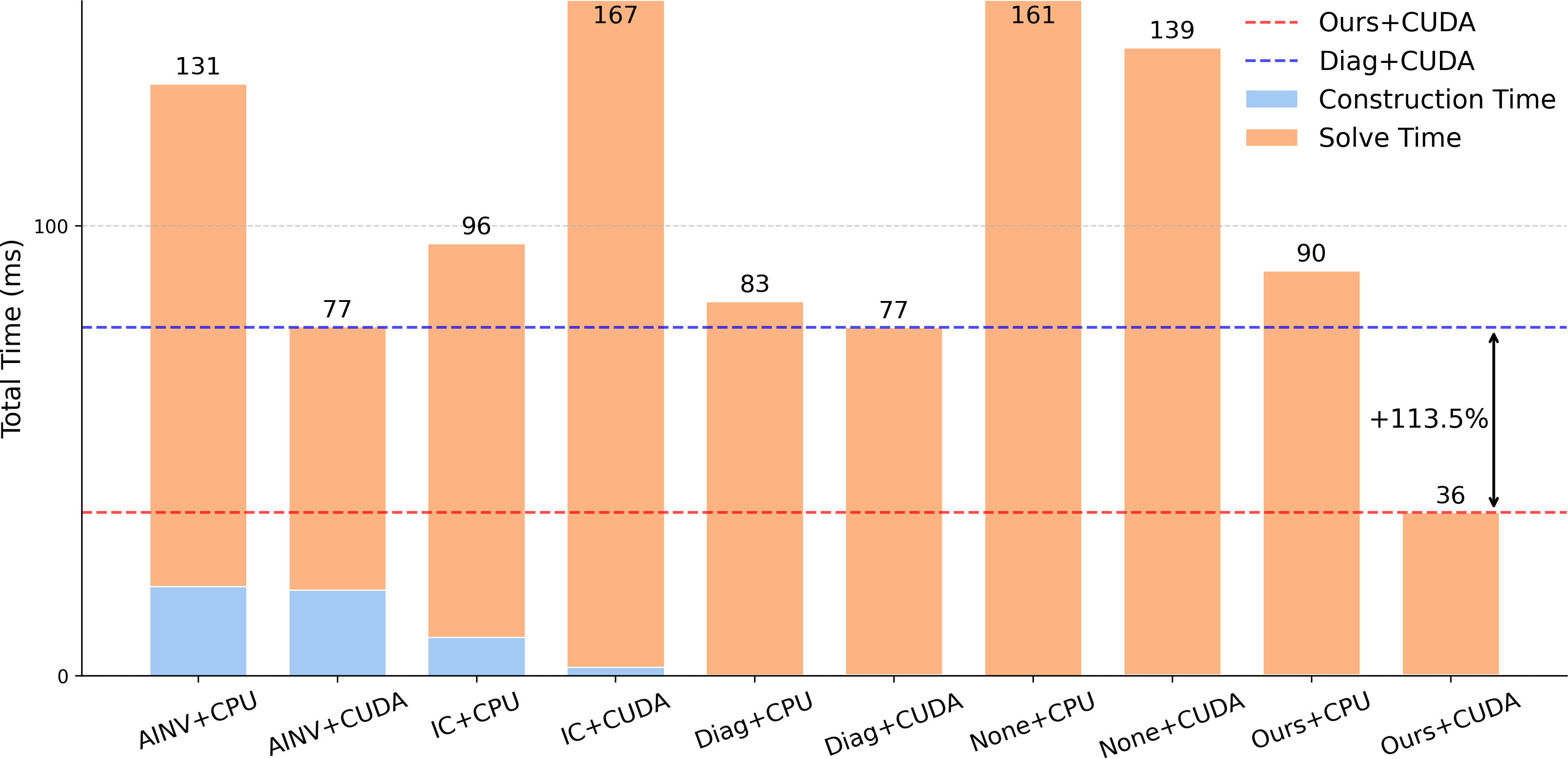}}
    \subfigure[Scalability]{\includegraphics[width=0.23\linewidth]{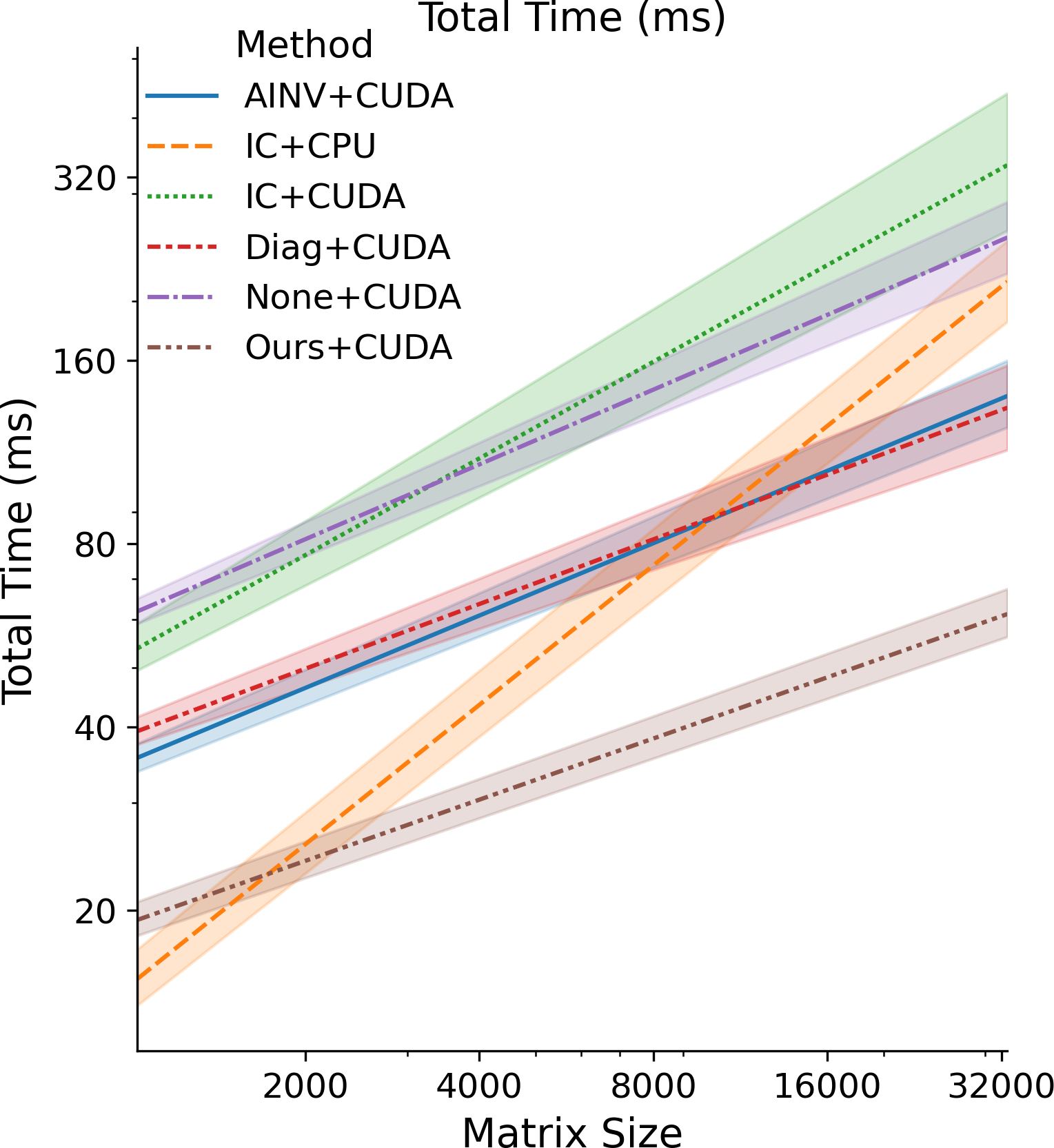}}
    \subfigure[Convergency]{\includegraphics[width=0.23\linewidth]{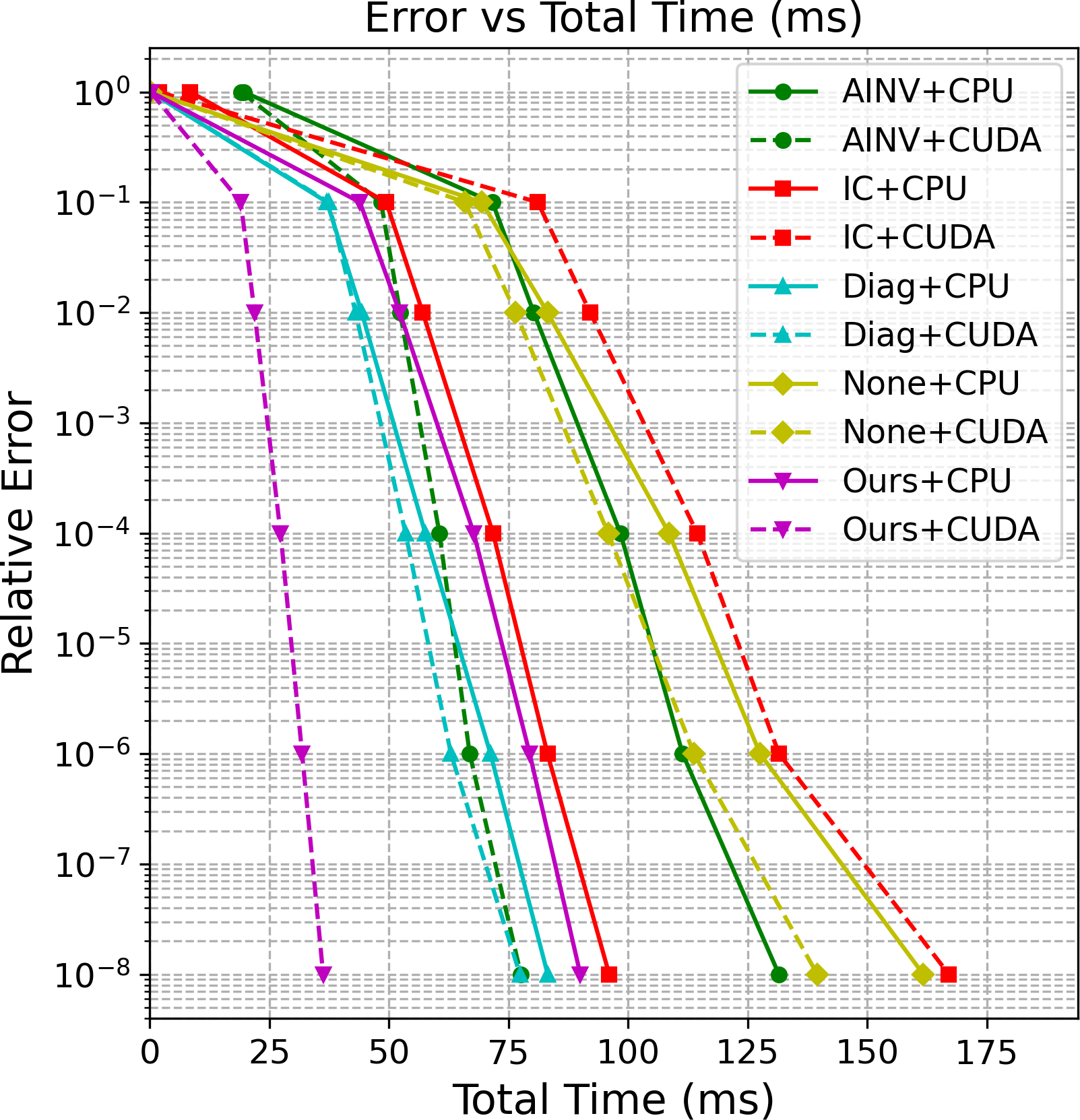}}
    \caption{{Performance of CG with different preconditioners for the heat problem.} 
    Figure (a) compares the average total solve time $T_\mathrm{total}$ and preconditioner's construction time $T_\mathrm{construct}$ of CG with different preconditioners and devices. Figure (b) illustrates the relationship between matrix size and $T_\mathrm{total}$, including its 95\% confidence interval, demonstrating the superior scalability of our approach on GPUs.
    Figure (c) compares the total solve time required to achieve different $\mathrm{rtol}$. }
    \label{fig:intro-comparsion}
\end{figure}

\paragraph{Performance Analysis Across Convergence Thresholds} Table \ref{tab:time-stats} summarizes the total solve time $T_{\mathrm{total}}$ and iteration numbers $k$ of CG solvers using different preconditioners up to various convergence thresholds. 
Although the Diagonal preconditioner has the smallest construction overhead, it results in a limited improvement in convergence rate.
Compared with previous preconditioners, our approach achieves the smallest $T_{\mathrm{total}}$ to the given $\mathrm{rtol}$.
We also compare the scalability of our approach with that of baseline methods in \cref{fig:intro-comparsion}, and our approach is still efficient as the matrix size increases.

\paragraph{Runtime Analysis of Routines in CG solvers} Table \ref{tab:time-decomposition} breaks down the runtime of each routine in CG solvers. Our approach reduces $T_{\mathrm{total}}$ by balancing three key factors: low setup cost (comparable to Diag) $T_{\mathrm{construct}}$, reasonable iteration counts $k$ (similar to IC), and GPU-efficient operations $T_{\mathrm{apply}}$. This trade-off makes our approach particularly competitive in GPU settings. 
Although IC achieves comparable iteration counts $k$, its computational cost for applying the preconditioner (the triangular solves) is substantially higher than ours on GPUs, resulting in degraded performance.

\begin{table}[htbp]
\centering
\caption{Comparison between different preconditioners for the heat problem. 
Total time $T_{\mathrm{total}}$ (ms), total iterations $k$ (in parentheses), and preconditioner's construction time $T_{\mathrm{construct}}$ (Cons.) are listed in the table. 
The best value is in bold, and a lower value indicates better performance.}
\label{tab:time-stats}
\begin{tabular}{c|cccc|cccc}
\toprule
    \multirow{2}{*}{Stage}  & \multicolumn{4}{|c|}{CPU} & \multicolumn{4}{c}{GPU}\\
    & Diag & IC & AINV  & Ours & Diag & IC & AINV  & Ours\\
\midrule
Cons. & \textbf{0.126} & 8.426 & 19.308 & 0.181 & 0.196 & 1.866 & 18.924 & 0.181 \\
$10^{-2}$ & 44(309) & 57(126) & 80(199) & 52(124) & 43(309) & 92(115) & 52(199) & \textbf{22}(124) \\
$10^{-4}$ & 58(383) & 72(157) & 98(246) & 68(154) & 53(384) & 114(143) & 61(246) & \textbf{27}(154) \\
$10^{-6}$ & 71(450) & 83(180) & 111(282) & 79(176) & 63(442) & 132(164) & 67(280) & \textbf{32}(176) \\
$10^{-8}$ & 83(511) & 96(205) & 132(328) & 90(197) & 77(520) & 167(204) & 78(330) & \textbf{36}(197) \\
\bottomrule
\end{tabular}
\end{table}

\begin{table}[htbp]
\centering
\caption{Time breakdown of each routine in preconditioned CG solvers on GPUs and CPUs for the heat problem. All timings are reported in milliseconds.}
\label{tab:time-decomposition}
\begin{tabular}{ccccccccc}
\toprule
 \multirow{2}{*}{Method} & \multicolumn{4}{c}{CPU} & \multicolumn{4}{c}{GPU} \\
 & $k$ & $T_{\mathrm{construct}}$ & $T_{\mathrm{apply}}$ & $T_{\mathrm{cg}}$ & $k$ &$T_{\mathrm{construct}}$ & $T_{\mathrm{apply}}$ & $T_{\mathrm{cg}}$\\
\midrule
Diag & 511 & 0.12 & 0.01 & \multirow{4}{*}{0.16} & 520 & 0.12 & 0.01 & \multirow{4}{*}{0.14} \\
IC   & 205 & 8.54 & 0.26 &      & 205 & 1.88 & 0.80 & \\
AINV & 328 & 19.0 & 0.17 &      & 330 & 19.0 & 0.03 & \\
Ours & 197 & 0.18 & 0.29 &      & 197 & 0.18 & 0.04 & \\
\bottomrule
\end{tabular}
\end{table}

\paragraph{Comparison to Learning-based Approaches}
We compare our method with existing learning-based approaches on the dataset provided by previous works in Table \ref{tab:benchmark-learning}. We adopt their open-source implementation with the default settings. Compared to previous learning-based methods, our approach demonstrates better performance particularly on GPUs.  The matrix size provided in \citet{li2023learning} is sufficiently small that Diagonal preconditioner on CPUs achieves the best performance, while our approach attains the second-best performance. Compared to \citet{hausner2024neural}, our approach achieves a lower total solve time ($T_{\mathrm{total}}$) on GPUs.

\begin{table}[h!]
\caption{Comparsion to previous works. Total time $T_{\mathrm{total}}$ (ms) and total iterations $k$ (in parentheses) until relative residual norm is less than $\mathrm{rtol}=10^{-8}$ are listed in the table.
Prev. corresponds to the performance of previous works.
The lowest value is in bold and the second lowest is underlined.}
\label{tab:benchmark-learning}
\centering
\begin{tabular}{ccccccc}
\toprule
    Device & Test Case  & Diag & IC & AINV & Prev. & Ours\\
\midrule
\multirow{2}{*}{CPU}
&\citet{li2023learning}      & \textbf{12}(208) & 19(90) & 33(108) & 26(108) & \underline{17}(102)\\
&\citet{hausner2024neural} & \textbf{799}(970) & 1387(438) & 3235(753) & \underline{1139}(354) & 1320(456)\\
\multirow{2}{*}{GPU}
&\citet{li2023learning} & \underline{29}(208) & 45(87) & 30(108) & 51(108) & \textbf{26}(102) \\
&\citet{hausner2024neural}  & \underline{166}(970) & 440(385) & 2456(753) & 1040(354) & \textbf{132}(456)\\
\bottomrule
\end{tabular}
\end{table}

\subsection{Generalizability and Robustness}\label{subsec:general}

We present the performance of our approach on out-of-distribution test samples in Table \ref{tab:generalization}.
For the heat problem, we evaluate at a fixed mesh density of 1e-3 (Heat-Density) and further assess generalization to finer meshes with more than 32k nodes (Heat-Large).
For the hyperelasticity problem, we test on the same geometry with a finer mesh (22,618 nodes). 
For the synthetic problem, we use matrices with size ranging from 48k to 96k (Synthetic-Large).
Comparing the relative promotions on each test case, our approach generalizes well to unseen resolutions and physical parameters.

\begin{table}[h!]
\centering
\caption{Test on out-of-distribution data on GPUs. The total time (ms) and total iterations (in parentheses) are reported. The best value is in bold, and a lower value indicates better performance. In-distribution (In) and out-of-distribution (Out) relative promotions are recorded.
}
\label{tab:generalization}
\begin{tabular}{cccccccc}
\toprule
    \multirow{2}{*}{Test Case} & \multirow{2}{*}{Diag}  & \multirow{2}{*}{IC}  & \multirow{2}{*}{AINV}   & \multirow{2}{*}{Ours}  & \multicolumn{2}{c}{Rel. Prom. ($\uparrow$)} \\
    &&&&& Out & In\\
\midrule
    Heat-Density& \underline{62}(468) & 135(175) & 67(307) & \textbf{35}(201) & 80\% & 113\% \\
    Heat-Large& \underline{251}(1033) & 808(388) & 407(740) & \textbf{154}(409) & 62\% & 73\% \\
    Hyperelasticity& \underline{326}(1005) & 667(287) & 1154(496) & \textbf{236}(359) & 72\% & 68\% \\
    Synthetic-Large& \underline{485}(1575) & 1616(781) & 19840(5591) & \textbf{347}(559) & 39\% & 75\%\\
\bottomrule
\end{tabular}
\end{table}

\subsection{Ablation Study on Loss Functions}\label{subsec:lossfunction}

We compare SAI loss with cosine similarity loss $\mathcal{L}_\mathrm{CS} = \mathbf{w}^\top \bA\bM^{-1}\mathbf{w} / \|\mathbf{w}\| \| \bA\bM^{-1}\mathbf{w} \|$ \citep{trifonov2025canmpgnnapproximate}
and the scale variant loss $\mathcal{L}_2= \|\bA\bM^{-1}\mathbf{w} - \mathbf{w}\|^2_2$. 
As shown in Table \ref{tab:lossfunction}, although $\mathcal{L}_{\mathrm{2}}$ performs similarly on the heat problem, the proposed SAI loss demonstrates better effectiveness on more difficult test cases.

\begin{table}[h!]
    \centering
    \caption{Comparison of different losses. The average iteration counts are recorded. Since $T_\mathrm{construct}$ and $T_\mathrm{apply}$ are shared, a lower value indicates a better result and a smaller total solving time.}
    \label{tab:lossfunction}
    \begin{tabular}{ccccc}
    \toprule
       Loss  & Heat & Poisson & Hyperelasticity & Synthetic \\
    \midrule
        $\mathcal{L}_2$ & \textbf{195.6} & 134.1 & 185.4  & 2109.8 \\
        $\mathcal{L}_{\mathrm{CS}}$ & 207.3 & 133.4 & 182.5 & 2185.7\\
        $\mathcal{L}_{\mathrm{SAI}}$ & 197.4 & \textbf{128.8} & \textbf{175.7} & \textbf{1122.0} \\
    \bottomrule
    \end{tabular}
\end{table}

\subsection{Condition Number of Preconditioned Matrix}\label{subsec:condition-number}

\begin{figure}[h!]
    \centering
    \subfigure[$\kappa$]{\includegraphics[width=0.48\linewidth]{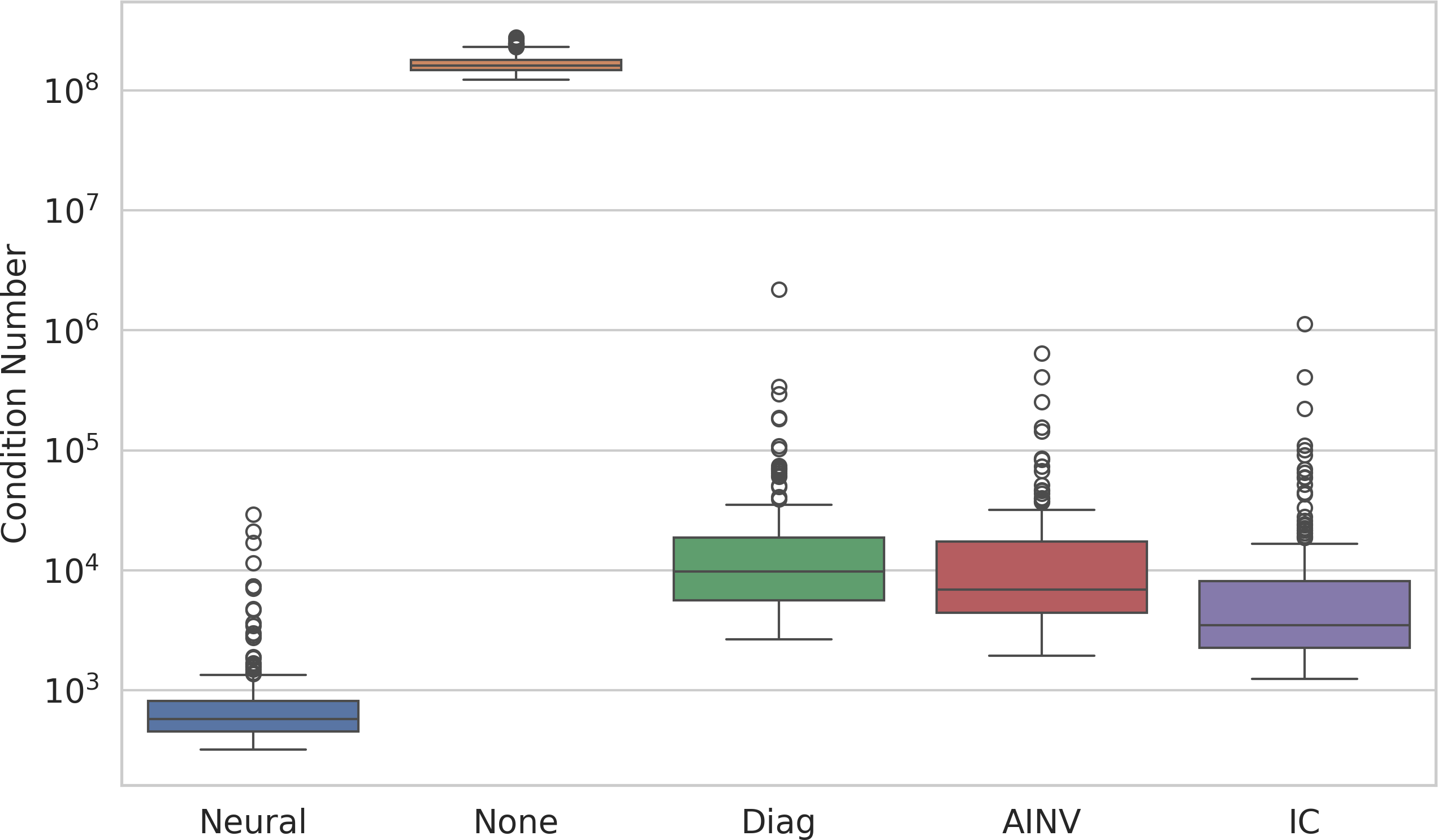}}
    \subfigure[$\kappa_{\mathrm{Kaporin}}$]{\includegraphics[width=0.48\linewidth]{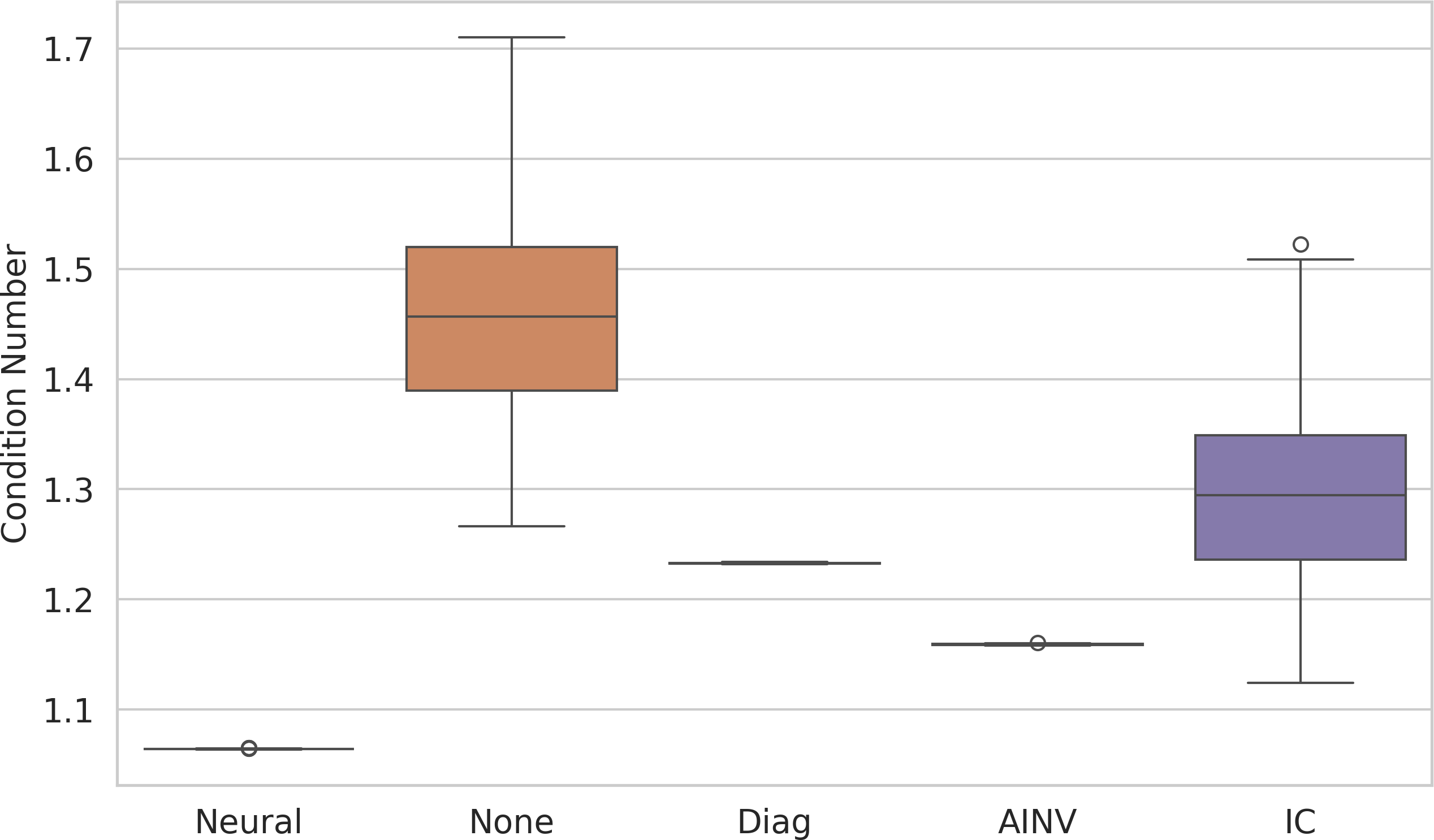}}
    \caption{Condition number distributions. Median, IQR, and outliers are shown. A smaller conditioner number indicates better performance of the preconditioner, and the lower bound is 1.}
    \label{fig:cond}
\end{figure}

The condition numbers provide a theoretical measure of the preconditioner's performance. Besides the standard condition number $\kappa$, we also compute the Kaporin's condition number \cite{kaporin1994new}, which is the ratio of the arithmetic mean to the geometric mean of the matrix eigenvalues $\lambda_i$:
\begin{equation}
   \kappa(\bA \bM^{-1}) = \frac{\max_i \lambda_i}{\min_i \lambda_i}, \quad \kappa_{\mathrm{Kaporin}}(\bA\bM^{-1}) = \frac{(\sum_{i=1}^{N} \lambda_i) / N }{(\lambda_1\lambda_2 \cdots \lambda_N)^{\frac{1}{N}}}.
\end{equation}
We evaluate the condition numbers on a simplified heat problem on a single mesh with 3,764 nodes and varying diffusivity coefficients.
Figure \ref{fig:cond} demonstrates that our method notably reduces the condition number when compared with earlier traditional approaches, revealing its efficacy.
\vspace{-5px}

\section{Conclusion}

This work proposes a learning-based approach for constructing SPAI preconditioners, aiming at accelerating the convergence of CG solvers on GPUs. Our method leverages the natural alignment between the local propagation mechanism of GNNs and the localized structure of SPAI preconditioners. Furthermore, we propose the SAI loss to reduce the computational cost of training data generation and improve the quality of the learned preconditioners. Experimental results demonstrate that our approach delivers consistent speedups over traditional and existing learning-based methods, with improved robustness, generalization, and compatibility with parallel computing architectures.

\paragraph{Limitations and Future Works}\label{para:limitations} Although our approach outperforms existing methods in accelerating CG solvers, several limitations remain. First, we enforce the sparsity pattern of $\bG$ to exactly match that of $\bA$, while many existing approaches employ dynamic dropping strategies to further limit fill-in or incorporate two-hop connection in $\bG$ to improve the effectiveness of preconditioner. Second, while our framework could potentially be extended to other Krylov subspace methods such as GMRES or integrated into multigrid frameworks (e.g., as a smoother), this work focuses specifically on SPD systems and the CG solver. Third, our current work is limited by the memory of a single GPU, and a promising direction for future work is to scale our approach to multi-GPU systems using techniques from distributed GNNs. 

\begin{ack}
This work is supported by the National Key R\&D Program of China (2022YFB3303400) and the National Natural Science Foundation of China (62025207). Tao Du acknowledges the research funding support from Tsinghua University and Shanghai Qi Zhi Institute.
\end{ack}

\bibliographystyle{unsrtnat}
\setcitestyle{square,numbers,comma}
{
    \small
    \bibliography{neurips_2025}
}

\appendix

\section{Preconditioned Conjugate Gradient Algorithm}\label{app:pcg}

We describe the standard PCG algorithm \cite{bar1994templates} here.

\begin{algorithm}
\caption{Preconditioned Conjugate Gradient} \label{alg:myalgorithm}
\begin{algorithmic}
\STATE $i \leftarrow 0$
\STATE $\mathbf{r} \leftarrow \mathbf{b} - A \mathbf{x}$
\STATE $\mathbf{d} \leftarrow M^{-1} \mathbf{r}$  \hfill \texttt{// Apply Preconditioner\hspace{3cm}}
\STATE $\delta_{\text{new}} \leftarrow \mathbf{r}^T \mathbf{d}$
\STATE $\delta_{0} \leftarrow \delta_{\text{new}}$
\WHILE{$i < i_{\text{max}}$ \textbf{and} $\delta_{\text{new}} > \varepsilon^2 \delta_{0}$}
    \STATE $\mathbf{q} \leftarrow A \mathbf{d}$
    \STATE $\alpha \leftarrow \frac{\delta_{\text{new}}}{\mathbf{d}^T \mathbf{q}}$
    \STATE $\mathbf{x} \leftarrow \mathbf{x} + \alpha \mathbf{d}$
    \IF{$i$ is divisible by 50}
        \STATE $\mathbf{r} \leftarrow \mathbf{b} - A \mathbf{x}$
    \ELSE
        \STATE $\mathbf{r} \leftarrow \mathbf{r} - \alpha \mathbf{q}$
        \STATE $\mathbf{s} \leftarrow M^{-1} \mathbf{r}$ \hfill \texttt{// Apply Preconditioner\hspace{3cm}}
        \STATE $\delta_{\text{old}} \leftarrow \delta_{\text{new}}$
        \STATE $\delta_{\text{new}} \leftarrow \mathbf{r}^T \mathbf{s}$
        \STATE $\beta \leftarrow \frac{\delta_{\text{new}}}{\delta_{\text{old}}}$
        \STATE $\mathbf{d} \leftarrow \mathbf{s} + \beta \mathbf{d}$
    \ENDIF
    \STATE $i \leftarrow i + 1$
\ENDWHILE
\end{algorithmic}
\end{algorithm}

\section{Proof of the Inequality}\label{app:proof-of-scal}

The inequality concerns the condition number $\kappa(\bA\bM^{-1})$.
The error matrix is defined in the main text as $\bE = \bA \bM^{-1} / \|\bA\|  - \mathbf{I}$.
From this definition, we can write:
\begin{equation}
    \mathbf{I} + \bE = \frac{\bA \bM^{-1}}{\|\bA\|}.
\end{equation}
The condition number $\kappa(\bA\bM^{-1})$ is given by:
\begin{equation}
    \kappa(\bA\bM^{-1}) = \frac{\sigma_{\max}(\bA\bM^{-1})}{\sigma_{\min}(\bA\bM^{-1})}.
\end{equation}
Since for any non-zero scalar $c$, $\kappa(c\mathbf{X}) = \kappa(\mathbf{X})$, we have:
\begin{equation}
    \kappa(\bA\bM^{-1}) = \kappa\left(\frac{\bA\bM^{-1}}{\|\bA\|}\right) = \kappa(\mathbf{I} + \bE) \frac{\sigma_{\max}(\mathbf{I} + \bE)}{\sigma_{\min}(\mathbf{I} + \bE)}.
\end{equation}

We use Weyl's inequality for singular values. For any square matrices $\mathbf{X}$ and $\mathbf{Y}$ of the same dimensions, Weyl's inequality states:
\begin{equation}
    |\sigma_k(\mathbf{X} + \mathbf{Y}) - \sigma_k(\mathbf{X})| \le \sigma_{\max}(\mathbf{Y}),
\end{equation}
where $\sigma_k(\cdot)$ is the $k$-th singular value and $\sigma_{\max}(\mathbf{Y})$ is the largest singular value of $\mathbf{Y}$ (i.e., its spectral norm $\|\mathbf{Y}\|_2$).

Let $\mathbf{X} = \mathbf{I}$ (the identity matrix) and $\mathbf{Y} = \bE$. The singular values of $\mathbf{I}$ are all $1$, so $\sigma_k(\mathbf{I}) = 1$ for all $k$.
Applying Weyl's inequality to $\mathbf{I} + \bE$:
\begin{equation}
    |\sigma_k(\mathbf{I} + \bE) - \sigma_k(\mathbf{I})| \le \sigma_{\max}(\bE)
\end{equation}
\begin{equation}
    |\sigma_k(\mathbf{I} + \bE) - 1| \le \sigma_{\max}(\bE).
\end{equation}
This implies that for any singular value $\sigma_k(\mathbf{I} + \bE)$:
\begin{equation}
    1 - \sigma_{\max}(\bE) \le \sigma_k(\mathbf{I} + \bE) \le 1 + \sigma_{\max}(\bE).
\end{equation}
This inequality holds for both the maximum and minimum singular values of $\mathbf{I} + \bE$:
\begin{align}
    \sigma_{\max}(\mathbf{I} + \bE) &\le 1 + \sigma_{\max}(\bE) \\
    \sigma_{\min}(\mathbf{I} + \bE) &\ge 1 - \sigma_{\max}(\bE).
\end{align}
For the lower bound on $\sigma_{\min}(\mathbf{I} + \bE)$ to be positive, and thus for $\mathbf{I} + \bE$ to be invertible, we require $\sigma_{\max}(\bE) < 1$. This condition is typically met when the preconditioner $\bM$ is effective, making $\bA\bM^{-1}/\|\bA\|$ close to $\mathbf{I}$.

Under the condition $\sigma_{\max}(\bE) < 1$, we can bound the condition number:
\begin{equation}
    \kappa(\bA\bM^{-1}) = \kappa(\mathbf{I} + \bE) = \frac{\sigma_{\max}(\mathbf{I} + \bE)}{\sigma_{\min}(\mathbf{I} + \bE)} \le \frac{1 + \sigma_{\max}(\bE)}{1 - \sigma_{\max}(\bE)}.
\end{equation}
This establishes the first part of the inequality.

For the approximation, if $\sigma_{\max}(\bE)$ is small (i.e., $\sigma_{\max}(\bE) \ll 1$), we can use the Taylor expansion for $(1-x)^{-1} = 1 + x + x^2 + \dots$ for $|x|<1$.
Let $x = \sigma_{\max}(\bE)$. Then:
\begin{equation}
    \frac{1 + x}{1 - x} = (1+x)(1-x)^{-1} = (1+x)(1 + x + x^2 + O(x^3)) = 1 + 2x + 2x^2 + O(x^3).
\end{equation}
For $x \ll 1$, we can approximate this as:
\begin{equation}
    \frac{1 + \sigma_{\max}(\bE)}{1 - \sigma_{\max}(\bE)} \approx 1 + 2\sigma_{\max}(\bE).
\end{equation}
Finally, recall that the spectral norm $\|\bE\|_2$ is defined as $\sigma_{\max}(\bE)$. Therefore,
\begin{equation}
    \kappa(\bA\bM^{-1}) \le \frac{1 + \sigma_{\max}(\bE)}{1 - \sigma_{\max}(\bE)} \approx 1 + 2\sigma_{\max}(\bE) = 1 + 2\|\bE\|_2.
\end{equation}
This completes the proof of the inequality chain presented in the main text.

\section{Experiment details}
\subsection{Implementation of Traditional Baselines}\label{app:baselines}

\textbf{Traditional Baselines} We implement all the preconditioned conjugate gradient solvers in C++ code with reduced function call overhead, and use \texttt{nanobind} \cite{nanobind} to generate its Python binding. Even one method can have different implementations on different devices, leading to different iteration counts.

\begin{table}[htbp]
\centering
\caption{Implementation of baseline preconditioners}
\label{tab:implementation}
\begin{tabular}{cccccc}
\toprule
   Device  &  Diag  & IC & SPAI & AMG \\
\midrule
   CPU     &  Eigen \cite{eigenweb} & Eigen \cite{eigenweb} & cusplibrary \cite{Cusp} & PyAMG \cite{pyamg2023}\\
   GPU     &  Custom & cuSPARSE & cusplibrary \cite{Cusp} & AMGX \cite{Naumov2015AmgXAL}\\
\bottomrule
\end{tabular}
\end{table}

All evaluations are performed on an AMD Ryzen 5 5600 CPU and an NVIDIA GeForce RTX 3060 GPU. The CPU frequency is fixed at 4.0 GHz and the OpenMP thread count is set to 4. The source code is compiled using GCC 14.2 and CUDA 12.8.

\subsection{Dataset Configurations}\label{app:dataset}

\textbf{Heat Equation/Poisson Equation}: The equations are discretized using the finite element method. We generate 9,147 samples on tetrahedral meshes from the TetWild dataset \cite{hu2018tetwild}, employing P1 spatial discretization. The input node features of the GNNs are the nodes' position and the density.

\paragraph{Hyperelasticity}: We adopt standard optimization based time integration with stable Neo-Hookean material model \cite{Kim2022DynamicDI}, $\nu = 0.4$, and density $\rho = 1.0$. The geometry of the beam is defined as $H\times W\times L = 1\times1\times4$, with Dirichlet boundary conditions applied to the left and right boundaries. All physical parameters are specified in SI units. The domain is discretized at varying resolutions using TetGen \cite{Si2015TetGenAD}. The input node features are the nodes' position and one-hot vector to indicate whether the node is on the Dirichlet boundary.

\paragraph{Synthetic}: For each sample, we first construct $\mathbf{P}$ with a specified sparsity of $3\times 10^{-4}$. To enforce the SPD property of $\bA$, we compute $\bA = \mathbf{PP}^{\top} + \varepsilon \mathbf{I}$, where $\varepsilon=$1e-4 is chosen for numerical stability. The resulting matrix $\bA$ has sparsity of 1.2e-3 approximately. The matrix size of $\mathbf{P}$ is randomly selected between 12,000 and 24,000. The input node feature $i$ is the $i$-th row average of matrix $\bA$.

\section{Additional Experiments}

\subsection{Performance on CPUs}

\begin{table}[htbp]
\caption{Benchmark results across different datasets. Total time $T_{\mathrm{total}}$ (ms) and total iterations $k$ (in parentheses) until relative residual norm is less than $\mathrm{rtol}=10^{-8}$ are listed in the table. Lower values indicate better performance. The lowest value is in bold, and the second lowest is underlined.}
\label{tab:benchmark-full}
\centering
\begin{tabular}{ccccccc}
\toprule
    Device & Test Case& Diag & IC & AINV & Ours\\
\midrule
\multirow{4}{*}{GPU} & Heat Equation& \underline{77}(520) & 167(204) & 78(330) & \textbf{36}(197) \\
&Poisson Equation& \underline{45}(320) & 101(128) & 58(217) & \textbf{26}(128) \\
&Hyperelasticity& \underline{86}(464) & 202(117) & 247(266) & \textbf{51}(175) \\
&Synthetic System& \underline{445}(2775) & 1399(1808) & 5024(10896) & \textbf{253}(1122) \\ 
\midrule
\multirow{4}{*}{CPU}&Heat Equation& \textbf{83}(511) & 96(205) & 132(328) & \underline{90}(197)  \\
&Poisson Equation& \textbf{53}(331) & 68(135) & 94(208) & \underline{59}(128) \\
&Hyperelasticity& \textbf{287}(464) & \underline{299}(160) & 453(266) & 361(175) \\
&Synthetic System& \textbf{950}(2776) & \underline{1064}(1221) & 14377(10906) & 1251(1122) \\ 
\bottomrule
\end{tabular}
\end{table}

\subsection{Comparison to AMG}\label{app:amg}

\textbf{Algebraic Multigrid (AMG)} is a powerful tool for solving large-scale linear systems on both CPU and GPU architectures by exploiting the hierarchical structure of graph nodes. AMG can further serve as a preconditioner for CG solvers. 
While our method adopts a significantly different approach compared to AMG,
we also include a comparison between AMG and our method on our datasets.

\textbf{Benchmark Results}: Our evaluation employs the default configurations listed in AMGX and PyAMG library for the evaluation across all datasets, while more precise settings could be applied to further improve its performance.
As listed in Table \ref{tab:amg-benchmark}, while effective for Heat/Poisson Equation problems, AMG can degrade the performance of CG solvers compared to other baseline preconditioners in more complex scenarios, due to challenges in selecting optimal parameters for smoothers and cycles.

\begin{table}[h!]
\centering
\caption{Benchmark result across different datasets. Total time $T_{\mathrm{total}}$ (ms) and total iterations $k$ (in parentheses) until relative residual norm is less than $\mathrm{rtol}=10^{-8}$ are listed in the table. We examine two configurations: (1) AMG-preconditioned CG (AMG+CG) and (2) standalone AMG. 
"/" indicates that the default settings does not to converge within 10 seconds.
}
\label{tab:amg-benchmark}
\begin{tabular}{ccccccccc}
\toprule
    \multirow{2}{*}{Case}  & \multicolumn{3}{c}{CPU} & \multicolumn{3}{c}{GPU}\\
    & AMG+CG & AMG & Ours & AMG+CG & AMG & Ours \\
\midrule
    Heat & 47(17) & 117(69) & 90(197) & 20(27)& \textbf{17}(18) & 36(197) \\
    Heat-Large & 491(22) & 1312(103) & 1086(409) & \textbf{41}(10) & 94(17) & 154(409) \\
    Poisson& 37(14) & 67(42) & 59(128) & \textbf{10}(9) & 38(30) & 26(128) \\
    Hyperelasticity & 785(91) & / & 361(175) & 286(128) & / & \textbf{51}(175) \\
    Synthetic & 6970(551) & / & 1251(1122) & / & / & \textbf{253}(1122) \\
\bottomrule
\end{tabular}
\end{table}

\subsection{Ablation Study on Matrix Norms}
\label{sec:ablation_norm}
To validate the design choice of the custom matrix norm within our SAI loss, we conduct a targeted ablation study. As discussed in \cref{subsec:sailoss}, our proposed mean norm was selected for its dimension-agnosticism, robustness to outliers, and low computational cost, offering key advantages over conventional norms such as the Frobenius and L1 norms.
In this experiment, we evaluate the impact of this choice on the Heat dataset. We replace our norm with several standard alternatives while keeping all other model components and hyperparameters identical. The results are summarized in \cref{tab:norm_ablation}. They clearly demonstrate the superiority of our proposed norm.  This represents a significant improvement over the Frobenius norm and the L1 norm. This quantitative evidence confirms that the specific properties of our chosen norm are not merely theoretical advantages but translate directly into tangible performance gains in the optimization process.
\begin{table}[h!]
    \centering
    \caption{Ablation study on the choice of matrix norm within the SAI loss. We report the average PCG iterations required for convergence on the ``Heat'' dataset.}
    \label{tab:norm_ablation}
    \begin{tabular}{lc}
        \toprule
        Norm Type & Iterations \\
        \midrule
        Frobenius Norm ($\Vert \cdot \Vert_F$) & 222 \\
        L1 Norm ($\Vert \cdot \Vert_1$) & 231 \\
        \textbf{Ours} & \textbf{197} \\
        \bottomrule
    \end{tabular}
\end{table}

\subsection{Sensitivity Analysis of the Hyper-parameter $\varepsilon$}
\label{app:epsilon_sensitivity}
This section addresses the sensitivity of our method's performance to the value of the hyper-parameter $\varepsilon$ in \eqref{eq:major}. To evaluate this, we conducted an ablation study on the Heat dataset by using different values of $\varepsilon$. \cref{tab:epsilon_sensitivity} summarizes the validation performance (measured in iterations) for the tested $\varepsilon$ values. The results demonstrate that our method is robust, with stable performance across three orders of magnitude ($3\times10^{-4}$ to $3\times10^{-2}$), indicating that $\varepsilon$ requires no careful per-instance tuning.
However, an excessively large value of $\varepsilon$ (e.g., $3\times10^{-1}$) leads to training failure. We hypothesize that such a strong regularization term over-constrains the preconditioner, causing it to deviate excessively from the intended structure and thus degrading performance.

\begin{table}[htbp]
    \centering
    \caption{Sensitivity analysis of the hyper-parameter $\varepsilon$ on the Heat dataset. Performance is measured by the number of iterations required on the validation set.}
    \label{tab:epsilon_sensitivity}
    \begin{tabular}{ccccc}
        \toprule
        $\varepsilon = 3\times10^{-5}$ & $\varepsilon = 3\times10^{-4}$ & $\varepsilon = 3\times10^{-3}$ & $\varepsilon = 3\times10^{-2}$ & $\varepsilon = 3\times10^{-1}$ \\
        \midrule
        222 & 208 & 197 & 205 & Training Failure \\
        \bottomrule
    \end{tabular}
\end{table}

\subsection{Generalization Analysis Against Learning-Based Baselines}
\label{app:generalization_analysis}
This section provides a detailed analysis to address the generalization capabilities of our method compared to other learning-based approaches.

\paragraph{Comparison with Neural PCG}
\label{subsec:neural_pcg_ood}
For this experiment, we follow the OOD setting from \cref{tab:generalization} in the main paper and compare against Neural PCG \citep{li2023learning}. Specifically, we decrease the density in the heat problem to an unseen value. \autoref{tab:ood_neural_pcg} presents the total time (ms) and iteration counts (in parentheses) for both in-distribution and OOD settings.
Our method not only outperforms Neural PCG in the in-distribution setting but also demonstrates significantly better generalization. While all methods experience performance degradation in the OOD setting, our method remains the most efficient.
\begin{table}[h!]
    \centering
    \caption{OOD generalization comparison with Neural PCG \citep{li2023learning}. Total time (ms) and total iterations $k$ (in parentheses) are reported. The best and second-best results are bolded and underlined, respectively.}
    \label{tab:ood_neural_pcg}
    \begin{tabular}{lccccc}
        \toprule
        Setting & IC & Diagonal & AINV & Neural PCG \citep{li2023learning} & Ours \\
        \midrule
        In-Distribution & 45(87) & \underline{29}(208) & 30(108) & 51(108) & \textbf{26(102)} \\
        Out-of-Distribution & 337(876) & 262(1956) & \underline{179}(1039) & 268(743) & \textbf{167(1022)} \\
        \bottomrule
    \end{tabular}
\end{table}

\paragraph{Comparison with Neural IF}
\label{subsec:neural_if_ood}
We further test generalization against Neural IF \citep{hausner2024neural} by increasing the sparsity of the synthesized matrix from $10^{-3}$ to $2\times10^{-3}$. The results are summarized in \autoref{tab:ood_neural_if}.
Again, our method exhibits superior performance and robustness. In the more challenging OOD setting, our approach achieves a remarkable speedup over Neural IF, confirming its strong generalization capability.
\begin{table}[h!]
    \centering
    \caption{OOD generalization comparison with Neural IF \citep{hausner2024neural}. The best and second-best results are bolded and underlined, respectively.}
    \label{tab:ood_neural_if}
    \begin{tabular}{lccccc}
        \toprule
        Setting & IC & Diagonal & AINV & Neural IF \citep{hausner2024neural} & Ours \\
        \midrule
        In-Distribution & 440(385) & \underline{166}(970) & 2456(753) & 1040(354) & \textbf{132}(456) \\
        Out-of-Distribution & 616(414) & \underline{207}(1040) & 3220(804) & 1222(375) & \textbf{168}(470) \\
        \bottomrule
    \end{tabular}
\end{table}

\newpage
\section*{NeurIPS Paper Checklist}

\begin{enumerate}

\item {\bf Claims}
    \item[] Question: Do the main claims made in the abstract and introduction accurately reflect the paper's contributions and scope?
    \item[] Answer: \answerYes{} 
    \item[] Justification: The main algorithm and theoretical claims are outlined and analyzed in Section \ref{sec:ourapproach}. The effectiveness of our method is validated and discussed in Section \ref{sec:experiments}.

\item {\bf Limitations}
    \item[] Question: Does the paper discuss the limitations of the work performed by the authors?
    \item[] Answer: \answerYes{}{} 
    \item[] Justification: The paper discusses the limitations of the proposed method in Section~\ref{para:limitations}, including the fixed sparsity pattern constraint and the focus on SPD systems and CG solvers.

\item {\bf Theory assumptions and proofs}
    \item[] Question: For each theoretical result, does the paper provide the full set of assumptions and a complete (and correct) proof?
    \item[] Answer: \answerYes{} 
    \item[] Justification: All theorems include clearly stated assumptions and complete proofs in Section \ref{sec:ourapproach}. Supporting lemmas and prior results are properly referenced.

    \item {\bf Experimental result reproducibility}
    \item[] Question: Does the paper fully disclose all the information needed to reproduce the main experimental results of the paper to the extent that it affects the main claims and/or conclusions of the paper (regardless of whether the code and data are provided or not)?
    \item[] Answer: \answerYes{} 
    \item[] Justification: Section \ref{subsec:exp-setup} provides detailed descriptions of the experimental setup, including model architecture, training procedure, datasets, and evaluation metrics.

\item {\bf Open access to data and code}
    \item[] Question: Does the paper provide open access to the data and code, with sufficient instructions to faithfully reproduce the main experimental results, as described in supplemental material?
    \item[] Answer: \answerYes{}{} 
    \item[] Justification: Yes, the code will be made publicly available upon publication, along with detailed instructions to reproduce all main experimental results.

\item {\bf Experimental setting/details}
    \item[] Question: Does the paper specify all the training and test details (e.g., data splits, hyperparameters, how they were chosen, type of optimizer, etc.) necessary to understand the results?
    \item[] Answer: \answerYes{} 
    \item[] Justification: Section \ref{subsec:exp-setup} provides detailed descriptions of the experimental setup, including model architecture, training procedure, datasets, and evaluation metrics.

\item {\bf Experiment statistical significance}
    \item[] Question: Does the paper report error bars suitably and correctly defined or other appropriate information about the statistical significance of the experiments?
    \item[] Answer: \answerYes{} 
    \item[] Justification: The paper reports error bars for all key experimental results.
    \item[] Guidelines:
    \begin{itemize}
        \item The answer NA means that the paper does not include experiments.
        \item The authors should answer "Yes" if the results are accompanied by error bars, confidence intervals, or statistical significance tests, at least for the experiments that support the main claims of the paper.
        \item The factors of variability that the error bars are capturing should be clearly stated (for example, train/test split, initialization, random drawing of some parameter, or overall run with given experimental conditions).
        \item The method for calculating the error bars should be explained (closed form formula, call to a library function, bootstrap, etc.)
        \item The assumptions made should be given (e.g., Normally distributed errors).
        \item It should be clear whether the error bar is the standard deviation or the standard error of the mean.
        \item It is OK to report 1-sigma error bars, but one should state it. The authors should preferably report a 2-sigma error bar than state that they have a 96\% CI, if the hypothesis of Normality of errors is not verified.
        \item For asymmetric distributions, the authors should be careful not to show in tables or figures symmetric error bars that would yield results that are out of range (e.g., negative error rates).
        \item If error bars are reported in tables or plots, The authors should explain in the text how they were calculated and reference the corresponding figures or tables in the text.
    \end{itemize}

\item {\bf Experiments compute resources}
    \item[] Question: For each experiment, does the paper provide sufficient information on the computer resources (type of compute workers, memory, time of execution) needed to reproduce the experiments?
    \item[] Answer: \answerYes{} 
    \item[] Justification: Section \ref{subsec:exp-setup} and Appendix \ref{app:baselines} provides detailed descriptions of our experiment environments.
    
\item {\bf Code of ethics}
    \item[] Question: Does the research conducted in the paper conform, in every respect, with the NeurIPS Code of Ethics \url{https://neurips.cc/public/EthicsGuidelines}?
    \item[] Answer: \answerYes{}{} 
    \item[] Justification: We confirm that this research complies with the NeurIPS Code of Ethics.

\item {\bf Broader impacts}
    \item[] Question: Does the paper discuss both potential positive societal impacts and negative societal impacts of the work performed?
    \item[] Answer: \answerNA{} 
    \item[] Justification: The work is primarily focused on improving numerical methods for solving linear systems and does not involve direct human interaction, personal data, or deployment in societal contexts. While the method may inspire future research and contribute to computational efficiency in scientific computing, no specific societal impacts are foreseeable at this stage.
    
\item {\bf Safeguards}
    \item[] Question: Does the paper describe safeguards that have been put in place for responsible release of data or models that have a high risk for misuse (e.g., pretrained language models, image generators, or scraped datasets)?
    \item[] Answer: \answerNA{} 
    \item[] Justification: The paper does not involve the release of models or datasets with high risk for misuse. The proposed method is a numerical solver technique and does not generate content, process personal data, or operate in safety-critical domains without human oversight. Therefore, no specific safeguards are required.

\item {\bf Licenses for existing assets}
    \item[] Question: Are the creators or original owners of assets (e.g., code, data, models), used in the paper, properly credited and are the license and terms of use explicitly mentioned and properly respected?
    \item[] Answer: \answerYes{} 
    \item[] Justification: Yes, as the paper uses existing software libraries including cuSPARSE, Eigen, cusplibrary, AMGX, and PyAMG. All libraries are used in compliance with their respective licenses, and proper citations to the corresponding projects and documentation are provided in the manuscript.

\item {\bf New assets}
    \item[] Question: Are new assets introduced in the paper well documented and is the documentation provided alongside the assets?
    \item[] Answer: \answerYes{} 
    \item[] Justification: The code is well-documented, including instructions for setup, training, and evaluation, along with descriptions of dependencies, limitations, and licensing information. Documentation and code will be made open-source upon publication.

\item {\bf Crowdsourcing and research with human subjects}
    \item[] Question: For crowdsourcing experiments and research with human subjects, does the paper include the full text of instructions given to participants and screenshots, if applicable, as well as details about compensation (if any)? 
    \item[] Answer: \answerNA{} 
    \item[] Justification: The paper does not involve crowdsourcing or human subjects. It focuses on algorithmic development and numerical experiments for solving sparse linear systems, with no data collected from human participants.

\item {\bf Institutional review board (IRB) approvals or equivalent for research with human subjects}
    \item[] Question: Does the paper describe potential risks incurred by study participants, whether such risks were disclosed to the subjects, and whether Institutional Review Board (IRB) approvals (or an equivalent approval/review based on the requirements of your country or institution) were obtained?
    \item[] Answer: \answerNA{}{} 
    \item[] Justification: The paper does not involve research with human subjects. The work is focused on algorithmic development and numerical experiments for sparse linear systems.

\item {\bf Declaration of LLM usage}
    \item[] Question: Does the paper describe the usage of LLMs if it is an important, original, or non-standard component of the core methods in this research? Note that if the LLM is used only for writing, editing, or formatting purposes and does not impact the core methodology, scientific rigorousness, or originality of the research, declaration is not required.
    \item[] Answer: \answerNA{} 
    \item[] Justification: The core methodology of this research does not involve the use of large language models (LLMs).

\end{enumerate}

\end{document}